\documentclass[11pt,runningheads]{llncs}
\usepackage{graphicx}
\usepackage{amsmath}
\usepackage{amssymb}
\usepackage[font=small]{caption}
\usepackage{subcaption}
\usepackage{multirow,tabularx}
\usepackage[hyphens]{url}
\usepackage{hyperref}
\usepackage{float}
\usepackage{xcolor}
\usepackage{algorithm}
\usepackage{algpseudocode}

\begin{document}
\title{Variable Assignment Invariant Neural Networks for Learning Logic Programs}
\titlerunning{$\delta$LFIT2}
\author{Yin Jun Phua\inst{1}\orcidID{0000-0003-1178-8238} \and
Katsumi Inoue\inst{2, 3,1}\orcidID{0000-0002-2717-9122}}
\authorrunning{Y.J. Phua et al.}
\institute{Tokyo Institute of Technology, Japan \email{phua@c.citech.ac.jp} \and
National Institute of Technology, Japan \email{inoue@nii.ac.jp} \and
The Graduate University for Advanced Studies SOKENDAI, Japan}
\maketitle
\begin{abstract}
Learning from interpretation transition (LFIT) is a framework for learning rules from observed state transitions. LFIT has been implemented in purely symbolic algorithms, but they are unable to deal with noise or generalize to unobserved transitions. Rule extraction based neural network methods suffer from overfitting, while more general implementation that categorize rules suffer from combinatorial explosion. In this paper, we introduce a technique to leverage variable permutation invariance inherent in symbolic domains. Our technique ensures that the permutation and the naming of the variables would not affect the results. We demonstrate the effectiveness and the scalability of this method with various experiments. Our code is publicly available at \url{https://github.com/phuayj/delta-lfit-2}.

\end{abstract}
\section{Introduction}

Dynamic systems exist in every aspect of our world. Understanding and being able to model such dynamic systems allow us to predict and even control the outcomes of such systems. Learning from Interpretation Transition (LFIT) \cite{LFIT2013} is a framework that allows automatic construction of a model in the form of logic programs, based solely on the state transitions observed from dynamic systems. LFIT has been largely implemented with symbolic algorithms \cite{ribeiro2021polynomial} that utilize logic operations. These algorithms are therefore interpretable and independently verifiable. The resulting model, a logic program that describes the dynamics of the system, is also interpretable. However, these symbolic algorithms treat all data as equally valid and thus are vulnerable to ambiguous, conflicting or noisy data. Symbolic LFIT algorithms also require all state transitions to be observable, and thus cannot generalize to or predict unobserved data.

Recent advances in deep learning and neural network represent a good opportunity to address the above issues. The field of combining neural network and symbolic algorithms, dubbed Neural-Symbolic AI (NSAI) provides an interesting avenue for addressing the issues above, while retaining the advantages of being interpretable and verifiable. With the advent of large language models (LLM), there have been renewed interests \cite{ji2023survey} in combining symbolic reasoning with LLMs \cite{messina2021transformer}. While these are promising directions for increasing reliability in LLMs, they are still far from being able to tackle the issue of being able to understand and verify the output of a neural network model.

In the field of understanding dynamic systems, NN-LFIT \cite{ilp-2016} proposes an extraction-based method that trains a neural network and extracts a logic program. DFOL \cite{gao2022learning} learns first-order rules also extracting symbolic knowledge from trained neural network. An advantage of these methods is the ability to retain interpretability while also capturing statistical characteristics of the dataset. However, these methods place limitations on the architecture of the neural network, making it difficult to utilize advancements in the deep learning field. For instance, application of these methods to the infamous attention mechanism is not straightforward.

$\delta$LFIT+ \cite{phua2021learning} has been proposed to take advantage of the strengths of neural network, while producing a model that is interpretable and verifiable. Based on the LFIT framework, $\delta$LFIT+ takes as input a set of state transitions and outputs the logic program that describes the state transitions. By not performing extractions and not placing any constraints on the neural network, $\delta$LFIT+ can utilize Set Transformer \cite{DBLP:journals/corr/abs-1810-00825} to exploit the invariance in the permutation of the inputs. Instead of using gradient descent as an equivalence of the learning in symbolic algorithms, $\delta$LFIT+ trains a model that learns to perform the learning. In a sense, it is a meta-learning model that produces a symbolic model as an output.

However, there are several limitations to $\delta$LFIT+. First, while $\delta$LFIT+ is invariant to the order of state transitions in the input, it is not invariant to the permutation of the variables within each state. Next, $\delta$LFIT+ employs an "output node reuse" strategy to address scalability issues in the model architecture, but it also leads to a loss function that is difficult to optimize.

\begin{figure}[t]
    \centering
    \includegraphics[width=\textwidth]{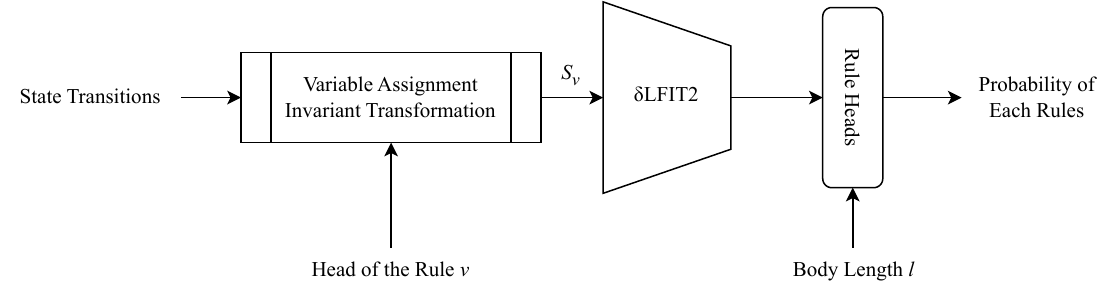}
    \caption{Overview of the $\delta$LFIT2 Framework.}
    \label{fig:delta-lfit-2}
\end{figure}

This paper proposes $\delta$LFIT2, shown in Figure \ref{fig:delta-lfit-2}, that addresses the above limitations. $\delta$LFIT2 introduces a variable assignment invariant technique that addresses the issue of variable permutation within states. $\delta$LFIT2 also utilizes multiple different output heads instead of reusing the same node for different purposes, leading to a smoother loss function.

\section{Related Work}

Recent advancements in deep learning models have renewed interests in NSAI. In particular, as LLMs gain attention and are used widely, the lack of reasoning ability \cite{valmeekam2022large} starts to become critical.

\vspace{-15pt}
\subsubsection{Extraction-based Methods}

CIL$^2$-P \cite{garcez2001symbolic} introduced the foundation for extracting symbolic knowledge from neural networks. While simple, the algorithm placed heavy constraints on the architecture of the neural network. Related to LFIT, NN-LFIT \cite{ilp-2016} proposed a method that trains a minimal neural network to model the dynamic system, then extracts symbolic rules based on the weights of the neural network.

\vspace{-15pt}
\subsubsection{Differentiable Programming-based Methods}

The Apperception Engine \cite{evans2021making}, $\delta$ILP \cite{evans2018learning} and Logic Tensor Network \cite{badreddine2022logic} proposed methods that leverage gradient descent to learn a matrix that can later be transformed into symbolic knowledge. These methods can be integrated with CNN or other neural network modules to further process continuous data. D-LFIT \cite{gao2022learning2} uses gradient descent to optimize a learnable matrix that semantically mirrors the $T_P$ operator. Compared to $\delta$LFIT2, these methods require optimization for every problem instance. Scalability issues also remain, as the size of the learnable matrix scales together with the problem size.

\vspace{-15pt}
\subsubsection{Integration-based Methods}

NeurASP \cite{DBLP:conf/ijcai/YangIL20}, Deepstochlog \cite{winters2022deepstochlog} and other similar works propose methods that either uses symbolic reasoning engines to drive neural network models or vice versa. Recent works that combine LLMs and reasoning such as \cite{messina2021transformer} work similarly.

\vspace{-15pt}
\subsubsection{Invariance in Deep Learning}

An increasing amount of work have started to focus on the permutation invariance inherent in the problem domain, leading to an increase in performance \cite{jumper2021highly} \cite{jaegle2021perceiver} \cite{SU2024127063}. While most works focus on spatial invariance, this work focuses primarily on the invariance within the semantics.

\vspace{3pt}

In contrast to the above works, the neural network in $\delta$LFIT2 performs the semantic equivalent of learning in the symbolic algorithm. Symbolic knowledge is produced directly by the output of the neural network, instead of extraction from the results of the gradient descent optimization.

\section{Background}

\subsubsection{Normal Logic Program for State Transitions}

A normal logic program for state transitions (NLP) $P$ is a set of logical state transition rules $R$ that are of the following form:
\begin{align}
    R: A(t+1) \leftarrow A_1(t) \wedge \dots \wedge A_m(t) \wedge
    \neg A_{m+1}(t) \wedge \dots \wedge \neg A_n(t)
    \label{eq:dynamic-normal-rule}
\end{align}
where $A(t+1)$ is the head of the rule $h(R)$ and everything to the right of $\leftarrow$ is known as the body $b(R)$. This represents a rule where $A(t+1)$ is true if and only if $b^+(R) = \{A_1(t), \dots, A_m(t)\}$ are true and $b^-(R) = \{A_{m+1}(t), \dots, A_n(t)\}$ are false. If we consider a system with a set of variables $\{A, A_1, \dots, A_m, A_{m+1}, \dots, A_n\}$, and $A(t)$ represents the state of the atom $A$ at time $t$, then the above is also a dynamic rule. Thus, the above rule can be described in plain English as, the state of the variable $A$ at time $t+1$ is true if and only if the state of $A_1, \dots, A_m$ is true and $A_{m+1}, \dots, A_n$ is false at time $t$.

Note however, that even though the rule has time arguments for each atom, $t + 1$ only appears in the head while $t$ only appears in the body. Therefore, we can also equally express it with the following propositional rule:
\begin{align*}
    A \leftarrow A_1 \wedge \dots \wedge A_m \wedge \neg A_{m+1} \wedge \dots \wedge \neg A_n
\end{align*}
In this case, $A$ can be any of $A_1, \dots, A_n$ and will not be considered as a cyclic rule because of the implicit time parameter attached to the atom.

\vspace{-10pt}

\subsubsection{LFIT}
\label{sec:lfit}

Given an NLP $P$ of such propositional rules, we can simulate the state transition of a dynamical system with the $T_P$ operator.

An Herbrand base $\mathcal{B}$ represents all variables involved within a dynamic system. An Herbrand interpretation $I$ is a subset of the $\mathcal{B}$ representing a snapshot of the state of the dynamic system. For an NLP $P$ and an Herbrand interpretation $I$, the immediate consequence operator (or $T_P$ operator) is the mapping ${T_P:2^{\mathcal{B}}
\rightarrow 2^{\mathcal{B}}}$:
\begin{align}
    T_P(I) = \{h(R)\ |\ R \in P, b^+(R) \subseteq I, b^-(R) \cap I =
    \emptyset \}.
\end{align}
which represents the prediction of the next state of the dynamic system given the model of it $P$.
Given a set of Herbrand interpretations $E$ and $\{T_P(I)\ |\ I \in E\}$, the LFIT framework outputs a logic program $P$ which completely represents the dynamics of $E$.

The LFIT framework can be described as an algorithm that takes a set of state transitions $S = \{(I, T_P(I))\ |\ I \in E\}$ and an initial NLP $P_0$ which is usually empty as inputs, then outputs an NLP $P$ such that $P$ is consistent with the input $S$. Various symbolic algorithms have been proposed to implement the LFIT framework. However, purely symbolic methods are unable to learn general rules where state transitions might be noisy or missing, thus NSAI methods have also been proposed.

\begin{figure}[t]
    \centering
    \includegraphics[width=\textwidth]{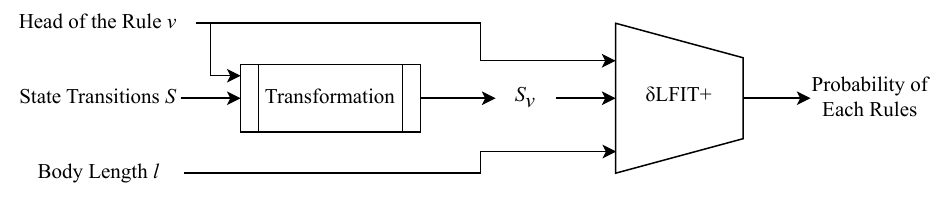}
    \caption{Overview of the $\delta$LFIT+ Framework.}
    \label{fig:delta-lfit-plus}
\end{figure}

\vspace{-15pt}

\subsubsection{$\delta$LFIT+}
\label{sec:delta-lfit-plus}

is a neural network implementation of the LFIT framework. Contrasting to many NSAI extraction based methods that extract symbolic knowledge from neural networks, $\delta$LFIT+ leverages neural networks to directly output symbolic knowledge, as depicted in Figure \ref{fig:delta-lfit-plus}.

$\delta$LFIT+ relies on the insight that the state transitions of a particular variable in a dynamic system can be uniquely described by a set of minimal rules. This means that a 1-to-1 mapping of state transitions to rules can be constructed. Indeed, this is the function approximated by the neural network.

$\delta$LFIT+ uses Set Transformer \cite{DBLP:journals/corr/abs-1810-00825} to process the state transitions in a permutation invariant manner. In the LFIT framework, state transitions are expressed as $S = \{(I, T_P(I))\ |\ I \in E\}$ where $E$ is every observable state. In $\delta$LFIT+, the rules are generated for a single variable at a time. Therefore, for a variable $v \in \mathcal{B}$ that we are generating the rules for, the set of state transitions are transformed to $S_v = \{(I, v \in T_P(I))\ |\ I \in E\}$ where the second element of the tuple is 1 if $v$ is in the next state and 0 otherwise. The elements are tokenized and fed into the Set Transformer.

To be able to output rules, all possible rules have to be enumerable. The number of possible rules can scale very quickly if no restrictions are applied. To make it feasible, only minimal rules according to the definition of \cite{phua2021learning}, which removes trivial rules and standardizes the order of the rule body, is considered. For every Herbrand base $\mathcal{B}$, a finite ordered set $\tau(\mathcal{B})$ that contains all minimal rules within $\mathcal{B}$ can be defined. $\delta$LFIT+ outputs rules by the head of the rule $v$ and by the length of the body $l$ one at a time. Since the output nodes of the neural network are fixed, $v$ and $l$ are provided as input which changes the interpretation of the outputs. Consider the Herbrand base $\mathcal{B} = \{p, q, r\}$. The rule for the output node at the first position can mean $p \wedge q$ if $l$ is 2 or $p \wedge q \wedge r$ if $l$ is 3. The head of the rule is also determined by the input $v$. Therefore, if the input for $v$ is given as $p$ and $l$ is 2, the rule for the output node at the first position is $p \leftarrow p \wedge q$.

\section{Proposed Method}

The loss function for $\delta$LFIT+ is not sufficiently smooth and thus very difficult to optimize. This is mainly caused by the non-continuous inputs of $l$ and $v$ that change the meaning of the output nodes. In particular, with the same set of state transitions, the neural network model needs to learn different sets of rules depending on $l$ and $v$. Another issue was the permutation of the variables in the dynamic system. While $\delta$LFIT+ utilized Set Transformer to ensure the invariance in the ordering of the input of state transitions, the ordering of the variables within the state itself still causes the model to output different results.

In this section, we describe our proposed method $\delta$LFIT2 which addresses the issues mentioned above.

\subsection{Variable Assignment Invariance}

\begin{figure}[t]
\centering
\begin{subfigure}{0.4\textwidth}
    \centering
    \includegraphics[width=0.6\textwidth]{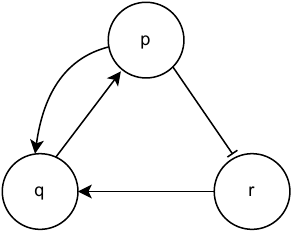}
    \caption{System 1}
    \label{fig:system-1}
\end{subfigure}
\begin{subfigure}{0.4\textwidth}
    \centering
    \includegraphics[width=0.6\textwidth]{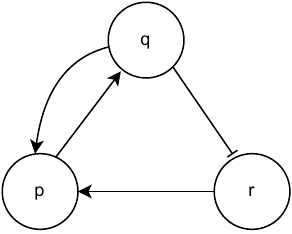}
    \caption{System 2}
    \label{fig:system-2}
\end{subfigure}
\caption{Two identical systems with the labels of the top nodes swapped. Arrows signify activation, whereas flat-edged symbols signify inhibition.}
\label{fig:two-identical-systems}
\end{figure}

In $\delta$LFIT+, the state transitions are tokenized based on the lexicographical ordering of the variable names. However, while variable names in real world systems are mostly based on their functions, names are ultimately arbitrary and unrelated to the actual dynamics. This means 2 topologically identical systems, with only variables being assigned differently, produce 2 different sets of state transitions. Having to learn both sets of transitions increases the complexity of the problem space. If instead, we can abstract the variable names (and thus the ordering of the variables), leading to just one set of state transitions, we can simplify the learning problem for the neural network model.

\begin{table}[t]
    \caption{Logic programs representing the two identical systems.}
    \label{table:logic-programs-identical}
    \centering
    \setlength{\tabcolsep}{12pt}
    \begin{tabular}{ c c }
    \hline
    System 1 & System 2 \\
    \hline
    $\begin{aligned}
        p &\leftarrow q. \\
        q &\leftarrow p \wedge r. \\
        r &\leftarrow \neg p.
    \end{aligned}$ & $\begin{aligned}
        q &\leftarrow p. \\
        p &\leftarrow q \wedge r. \\
        r &\leftarrow \neg q.
    \end{aligned}$ \\
    \hline
    \end{tabular}
\end{table}

Consider the identical systems shown in Figure \ref{fig:two-identical-systems}. These are boolean networks but can be represented by the logic programs in Table \ref{table:logic-programs-identical} respectively. Looking at the node on the top, which is variable $p$ in Figure \ref{fig:system-1} but variable $q$ in Figure \ref{fig:system-2}, they should result in the same identical set of rules, which is that they are activated by the variable in the bottom left. This is best achieved if the inputs into the model are the same.

The state transition that can be obtained for $p$ in System 1 and $q$ in System 2 respectively, based on the encoding described in Section \ref{sec:delta-lfit-plus}, is as follows:
\begin{align}
    S^1_p = \{(pqr, 1), (pq, 1), (p, 0), (\epsilon, 0), (r, 0), (qr, 1), (pr, 0), (q, 1)\} \\
    S^2_q = \{(pqr, 1), (pq, 1), (q, 0), (\epsilon, 0), (r, 0), (pr, 1), (qr, 0), (p, 1)\}
    \label{eq:system-1-transitions}
\end{align}
here we denote $S^1$ as state transitions obtained from System 1. Note that $S^1_p$ and $S^2_q$ are different.

To overcome this, we propose to use a permutation function $\Omega$ which reorders the variables, such that the variable that we are currently concerned about is in the first position.

As an example, consider the system in Figure \ref{fig:system-1} again. Since the variable that we are currently concerned is $p$, we can define the permutation function such that $\Omega_1: \{p, q, r\} \rightarrow \{v_0, v_1, v_2\}$ which maps $p$ to $v_0$, and so on. In System 2 on the other hand, when we are concerned with $q$, we can define a permutation function $\Omega_2: \{q, p, r\} \rightarrow \{v_0, v_1, v_2\}$ which maps $q$ to $v_0$ and so on. $S^2_q$ in equation (\ref{eq:system-1-transitions}). With both permutations, the transitions in equation (\ref{eq:system-1-transitions}) is transformed to:
{\small
\begin{align}
    S^1_p = \{(v_0 v_1 v_2, 1), (v_0 v_1, 1), (v_0, 0), (\epsilon, 0), (v_2, 0), (v_1 v_2, 1), (v_0 v_2, 0), (v_1, 1)\} \\
    S^2_q = \{(v_0 v_1 v_2, 1), (v_0 v_1, 1), (v_0, 0), (\epsilon, 0), (v_2, 0), (v_1 v_2, 1), (v_0 v_2, 0), (v_1, 1)\}
    \label{eq:system-1-transitions-transformed}
\end{align}
}
Notice that both transitions are now equivalent.

Both of these transitions lead to the same rule:
\begin{align*}
    v_0 \leftarrow v_1.
\end{align*}
Incidentally, this rule corresponds to the first output node when $l = 1$, according to the rule indexing in $\delta$LFIT+ \cite{phua2021learning}. This means that given the same input of (\ref{eq:system-1-transitions-transformed}), the neural network model only has to learn 1 output.

By applying the inverse permutation function $\Omega^{-1}_1$ and $\Omega^{-1}_2$ respectively, we can recover the following rules for System 1 and System 2 respectively:
\begin{align*}
    \text{System 1:}\ p &\leftarrow q.  \\
    \text{System 2:}\ q &\leftarrow p.
\end{align*}

In the example and in our implementation, we've used a rotating permutation, which rotates the variables in order. In practice, any permutation function can be used as long as it can be consistently applied to all variables and the inverse is as easily computable.

\subsection{Dynamic Rule Heads}
In $\delta$LFIT+, an "output node reuse" technique was employed to address the scalability issue. The naive way is to assign an output node for all possible minimal rules, which leads to $3^n$ number of nodes. This grows very quickly as $n$ gets larger. To address this, the rules are partitioned by body length and the neural network is constructed such that it covers the body length that consists of the most number of rules. For example, for Herbrand base with 3 variables, the body length that consists of the most number of rules is 2 with 12 such rules. The corresponding rules and the position of output nodes assigned is shown in table \ref{table:lfit_plus_rule}. Some output nodes utilized more (node 0 is used 4 times) and some less (nodes 8-11 are only used once), leading to an imbalance in the training data.

In $\delta$LFIT2, we construct separate output layers for each of the body lengths and dynamically load them into memory when required. This means that while the other layers are shared between each body length, the final layer is constructed separately. By dynamically loading and unloading the final layer, the memory usage of the model will not increase, even compared to that of $\delta$LFIT+.

In addition to each of the rules, the final layer also includes one extra output node, which indicates that there is no rule applicable for that body length. This is to combat the sparsity issue when there are no rules applicable for a specific body length. For example, the systems in Figure \ref{fig:two-identical-systems} has no applicable rules for $l = 3$. This sparsity issue makes optimization particularly difficult. In $\delta$LFIT+, this was countered by defining a subsumption matrix, where all subsumed rules are also given a small value. However, this subsumption matrix is expensive to compute and can grow very quickly as the number of variables increase.

\subsection{Overview of $\delta$LFIT2}
\label{sec:overview-delta-lfit-2}

\begin{table}[t]
\caption{Mapping of output nodes to rule bodies with various lengths.}
\label{table:lfit_plus_rule}
\centering
\tiny
\begin{tabular}{ c c c c c c c c c c c c }
\hline
Nodes & 0 & 1 & 2 & 3 & 4 & 5 \\
\hline
$l = 0$ & $\bot$ & - & - & - & - & - \\
$l = 1$ & $v_0$ & $v_1$ & $v_2$ & $\neg v_0$ & $\neg v_1$ & $\neg v_2$ \\
$l = 2$ & $v_0 \wedge v_1$ & $v_0 \wedge v_2$ & $v_1 \wedge v_2$ & $\neg v_0 \wedge v_1$ & $\neg v_0 \wedge v_2$ & $v_0 \wedge \neg v_1$ \\
$l = 3$ & $v_0 \wedge v_1 \wedge v_2$ & $\neg v_0 \wedge v_1 \wedge v_2$ & $v_0 \wedge \neg v_1 \wedge v_2$ & $\neg v_0 \wedge \neg v_1 \wedge v_2$ & $v_0 \wedge v_1 \wedge \neg v_2$ & $\neg v_0 \wedge v_1 \wedge \neg v_2$ \\
\hline
\hline
Nodes & 6 & 7 & 8 & 9 & 10 & 11 \\
\hline
$l = 0$ & - & - & - & - & - & - \\
$l = 1$ & - & - & - & - & - & - \\
$l = 2$ & $\neg v_1 \wedge v_2$ & $v_0 \wedge \neg v_2$ & $v_1 \wedge \neg v_2$ & $\neg v_0 \wedge \neg v_1$ & $\neg v_0 \wedge \neg v_2$ & $\neg v_1 \wedge \neg v_2$ \\
$l = 3$ & $v_0 \wedge \neg v_1 \wedge \neg v_2$ & $\neg v_0 \wedge \neg v_1 \wedge \neg v_2$ & - & - & - & - \\
\hline
\end{tabular}
\end{table}

The overall of $\delta$LFIT2 from the input of state transitions to the output of the probability of each rule is depicted in Figure \ref{fig:delta-lfit-2}. Contrasting to $\delta$LFIT+ in Figure \ref{fig:delta-lfit-plus}, we can see that the core neural network model now only has to be concerned with 1 input, which is the transformed set of state transitions $S_v$.

The model architecture of $\delta$LFIT2 consists of an embedding layer, which converts the tokens of the elements of $S_v$, into a sequence of embedding vectors. The sequence of embedding vectors is then fed into the Set Transformer, which produces a latent vector. We utilize both the encoder and the decoder part of the Set Transformer. The latent vector is then fed through several feed forward layers and finally, the result is fed to the final layer which then produces the probability of each rule being present or not.

In summary, $\delta$LFIT2 takes as input the head of the rule $v$, the body length $l$ and the observed state transitions $S$, then outputs a set of rules $R$ of body length $l$ and has $v$ as the head that partly explains $S$. So do we have to decide $v$ and $l$? The answer is that we do not. We would iterate $v$ over the entire Herbrand base and $l$ from 0 to the maximum body length, which is the number of variables in the Herbrand base. The rules predicted for each $v$ and $l$ are then specialized if they subsume each other and then combined into a final set of rules $P$ which is then the logic program that explains $S$.

To train $\delta$LFIT2, random state transitions are generated and rules that explain them are produced by the symbolic LFIT algorithm. We consider the synchronous semantic \cite{ribeiro2022learning} of the LFIT framework, where in a specific dynamic system, each state can only transition to another state deterministically. This means that we can generate many different systems by randomly generating pairs of interpretations. Once we enumerate every possible state and their random transitions, we can apply the symbolic LFIT algorithm \cite{ribeiro2015learning} and generate the corresponding rules. We can then split them up for every variable and for different rule body lengths to construct the training dataset.

\subsection{Scaling to Larger Systems}

Due to having to enumerate all possible rules at the output, $\delta$LFIT2 suffers from scalability issues in the architecture and the spatial dimension. This is in contrast to symbolic methods, where the scalability issue is in the time dimension.

A $\delta$LFIT2 model is trained for a fixed number of variables. This means that a model trained on 3 variables cannot be directly applied to learn a system with 5 variables that may have a rule with a body length of 5. However, especially in real world scenarios, rules that involve a large number of variables are incredibly rare. This is because a rule that involves many variables will remain inactive for most of the time. In the PyBoolNet repository \cite{10.1093/bioinformatics/btw682}, even a large system such as grieco\_mapk, a real world system from \cite{grieco2013integrative}, that has 53 variables, the longest rule only involves 10 variables.

To apply $\delta$LFIT2 to a larger system, consider the subsets of the Herbrand base of size $n$, where $n$ is the number of variables that the $\delta$LFIT2 model is trained for. For each of these subsets, we can define a mapping function from the subset to an index. This index is then used as the variable name for the $\delta$LFIT2 model. Once we obtain the rules, we can use the inverse of the mapping function to map from the index to the original variables.

By using the assumptions of the synchronous semantics, where every state deterministically only transitions to one state, it is possible to discard subsets of variables that produce transitions to different states. If all subsets for a particular variable are discarded, we know that the body length must be larger than the length of the subset. This can be used to determine whether a $\delta$LFIT2 model trained on a larger system is required when the maximum body length is unknown.

\section{Experiments}

To verify the effectiveness of the above proposed improvements, we report the results of the experiments on various dynamic systems in this section.

\vspace{-15pt}
\subsubsection{Datasets}

The boolean networks used in the experiments are taken from the PyBoolNet repository \cite{10.1093/bioinformatics/btw682}. The boolean networks in the repository contain both synthetic systems and real world systems. The boolean networks are converted into logic programs, and state transitions are obtained by applying the $T_P$ operator on all possible states $2^\mathcal{B}$. The training dataset is randomly generated as described in Section \ref{sec:overview-delta-lfit-2}.

\vspace{-15pt}
\subsubsection{Experimental Methods}

We use the Mean Squared Error (MSE) as the evaluation metric. Given the original program $P$ and the predicted program $P^\prime$, every possible interpretation of the Herbrand base $2^{\mathcal{B}}$, the MSE is calculated by the difference between the next states provided by the original program $\{T_P(I)\ |\ I \in 2^{\mathcal{B}}\}$ and the next states provided by the predicted program $\{T_{P^\prime}(I)\ |\ I \in 2^{\mathcal{B}}\}$. This metric is chosen because 2 different programs can generate the same exact state transitions. Therefore, a direct comparison of the rules will not reflect that both programs are equally valid semantically. All results presented are averaged across 3 runs.

We compare our proposed method $\delta$LFIT2 with NN-LFIT \cite{ilp-2016}, $\delta$LFIT \cite{DBLP:conf/ilp/PhuaI19} and $\delta$LFIT+ \cite{phua2021learning}. We denote the number of variables $n$ that $\delta$LFIT2 is trained on by a superscript $\delta$LFIT2$^n$.

\vspace{-15pt}
\subsubsection{Results}

\begin{table}[t]
\caption{MSE between state transitions generated by predicted logic programs for each method and state transitions from the original boolean network. $n$ denotes number of variables in the system, $b$ denotes the maximum rule body length.}
\label{table:experimental-results}
\centering
\begin{tabular}{l c c | c c c | c c}
    \hline
    Boolean Network & $n$ & $b$ & NN-LFIT & $\delta$LFIT & $\delta$LFIT+ & $\delta$LFIT2$^3$ & $\delta$LFIT2$^4$ \\
    \hline
    3-node a & 3 & 2 & \textbf{0.000} & 0.095 & 0.271 & \textbf{0.000} & 0.054 \\
    3-node b & 3 & 2 & 0.042 & 0.188 & 0.083 & \textbf{0.000} & \textbf{0.000} \\
    Raf & 3 & 2 & 0.333 & 0.253 & 0.188 & \textbf{0.000} & 0.025 \\
    5-node & 5 & 4 & 0.514 & \textbf{0.142} & 0.278 & 0.238 & 0.214 \\
    7-node & 7 & 3 & 0.299 & - & 0.223 & \textbf{0.035} & 0.152 \\
    WNT5A \cite{xiao2007impact} & 7 & 2 & 0.063 & - & 0.194 & \textbf{0.009} & 0.073  \\
    Circadian \cite{faure2006dynamical} & 10 & 4 & 0.260 & - & - & \textbf{0.033} & 0.136 \\
    Gene Network \cite{tournier2009uncovering} & 12 & 3 & 0.029 & - & - & \textbf{0.005} & 0.159 \\
    Budding Yeast \cite{irons2009logical} & 18 & 4 & - & - & - & \textbf{0.121} & 0.307 \\
    \hline
\end{tabular}
\end{table}

\begin{table}[t]
\caption{Comparison of results when only partial state transitions is given.}
\label{tab:partial-state-transitions}
\fontsize{8pt}{10pt}\selectfont
\centering
\begin{tabular}{c | c c c | c c c | c c c}
    \hline
     \multirow{2}{*}{Given} & \multicolumn{3}{c |}{3-node (a)} & \multicolumn{3}{c |}{3-node (b)} & \multicolumn{3}{c}{Raf} \\
     \cline{2-10}
     & NN-LFIT & $\delta$LFIT+ & $\delta$LFIT2$^3$ & NN-LFIT & $\delta$LFIT+ & $\delta$LFIT2$^3$ & NN-LFIT & $\delta$LFIT+ & $\delta$LFIT2$^3$ \\
     \hline
     3/8 & 0.542 & 0.319 & \textbf{0.308} & \textbf{0.292} & 0.472 & 0.362 & 0.458 & \textbf{0.319} & 0.358 \\
     4/8 & 0.500 & \textbf{0.264} & 0.296 & 0.403 & 0.403 & \textbf{0.321} & 0.417 & \textbf{0.264} & 0.287 \\
     5/8 & 0.333 & 0.389 & \textbf{0.188} & 0.292 & 0.389 & \textbf{0.267} & 0.458 & 0.306 & \textbf{0.242} \\
     6/8 & 0.375 & 0.236 & \textbf{0.183} & 0.167 & 0.222 & \textbf{0.154} & 0.333 & 0.278 & \textbf{0.150} \\
     7/8 & 0.083 & 0.292 & \textbf{0.071} & 0.083 & 0.195 & \textbf{0.079} & 0.250 & 0.236 & \textbf{0.063} \\
     \hline
\end{tabular}
\end{table}

The experimental results are shown in Table \ref{table:experimental-results}. We trained a 3-variable model noted as $\delta$LFIT2$^3$. We have also performed experiments with a model trained on 4 variables noted as $\delta$LFIT2$^4$, however due to time and resource constraints we were not able to fully train the model, leading to lower performance overall. In particular, while the entire input space for the 3-variable model is $(2^3)^{(2^3)} \approx 1.6 \times 10^7$ which can be generated in a trivial amount of time, the input space for the 4-variable model is $(2^4)^{(2^4)} \approx 1.8 \times 10^{19}$, which is difficult to generate. Nevertheless, we report the results in Table \ref{table:experimental-results}.

$\delta$LFIT2$^3$ achieved the best results in most cases. In particular, Raf which transitions to the same exact state from various states, NN-LFIT has a tendency to overfit and thus unable to produce the proper rules. In contrast, $\delta$LFIT2 is resistant to overfitting and thus able to recover the rules without issues. This highlights the advantage of our approach compared to extraction-based approaches.

Even for boolean networks that have a larger number of variables than what the $\delta$LFIT2 model were trained for, $\delta$LFIT2 can still recover a good amount of the rules. In particular, in WNT5A and Gene Network which only has rules with a maximum body length of 3, $\delta$LFIT2$^3$ is able to almost recover all the rules. For the 5-node network, a 4-body rule is included and therefore $\delta$LFIT2$^3$ was not able to recover the rules. $\delta$LFIT2$^4$ achieved better results than $\delta$LFIT2$^3$, but still fell short of $\delta$LFIT. We speculate that a fully trained $\delta$LFIT2$^4$ model would achieve better performance. Notably, we've demonstrated that $\delta$LFIT2 is able to scale to systems as large as 18 variables, which has $2^{18} = 262,144$ state transitions.

$\delta$LFIT2 was also able to produce rules that are more succinct and minimal than NN-LFIT. This is because $\delta$LFIT2 is trained specifically on training data that is produced by a symbolic algorithm that learns minimal rules. This is particularly evident for larger systems where NN-LFIT would produce large amount of rules, whereas $\delta$LFIT2 almost reconstructed the original minimal rules.

In terms of run time, once a $\delta$LFIT2 model is trained, the process of obtaining the rules only consists of forward inference. While the training process can take longer, it opens the avenue for a foundational model. In contrast, NN-LFIT has to perform the train, inference, and extraction process for every new system. And the extraction process in particular, can take a long period of time. For Budding Yeast, although NN-LFIT was able to train in a short amount of time, the production of the learned rules took more than a month, and thus we were not able to obtain a result.

We also performed experiments where a certain number of transitions is held out from the model. The results are shown in Table \ref{tab:partial-state-transitions}. At less than 50\% of the transitions provided, it is difficult for $\delta$LFIT2 to determine the correct rules. However, once exceeding 50\%, $\delta$LFIT2 can recover the rules better than other compared methods.

\section{Conclusion}

In this paper, we proposed a technique that leverages another invariance in the symbolic domain, namely the variable assignment invariance, and showed its effectiveness. We have also improved the architecture of the neural network model to make it easier to optimize. We combined these improvements and proposed a new method named $\delta$LFIT2 which greatly improves upon prior methods. We also showed that it is possible to apply $\delta$LFIT2 to a larger dynamic system. Future work can consider extending $\delta$LFIT2 to various asynchronous semantics and dynamic systems with delay or memory. Extending the $\delta$LFIT2 model to first order logic also represents an interesting avenue to explore. We hope this work will inspire more Neuro-Symbolic works that employ techniques to exploit variable name invariance in semantics.

\subsubsection*{Acknowledgements}

This work was supported by JSPS KAKENHI Grant Number JP22K21302, JP21H04905, the JST CREST Grant Number JPMJCR22D3. This research was also supported by ROIS NII Open Collaborative Research 2024-24S1203.

\appendix

\section{Implementation Details}
\label{app:implementation-details}

The model was implemented in PyTorch 2.1 with Python 3.10. The model architectures are as follows. All models include 3 components, i.e., the Set Transformer encoder, the Set Transformer decoder and a feed-forward layer. The activation function used is ReLU.

\subsection{$\delta$LFIT2$^3$}

We did not do any grid search for the following hyperparameters because we were happy with the performance. A smaller model can possibly be constructed.

\begin{itemize}
    \item Set Transformer Encoder
    \begin{itemize}
        \item Input dimensions: 256
        \item Output dimensions: 256
        \item Layers: 3
        \item Number of heads: 8
        \item Number of indices: 64
    \end{itemize}
    \item Set Transformer Decoder
    \begin{itemize}
        \item Input dimensions: 256
        \item Hidden dimensions: 256
        \item Output dimensions: 256
        \item Layers: 1
        \item Number of heads: 8
    \end{itemize}
    \item Feed-forward Layer
    \begin{itemize}
        \item Hidden dimensions: 1024
        \item Layers: 3
    \end{itemize}
    \item Learning rate: $1 \times 10^{-4}$
    \item Weight decay: $1 \times 10^{-4}$
    \item Optimizer: SGD
    \item Dropout: 0.2
    \item Layer normalization: yes
    \item Training data percentage: 95\%
    \item Test data percentage: 0.5\%
    \item Validation data percentage: 4.5\%
\end{itemize}

\subsection{$\delta$LFIT2$^4$}

We did not do any grid search for the following hyperparameters due to the training time required. A more optimized hyperparameter possibly still exists.

\begin{itemize}
    \item Set Transformer Encoder
    \begin{itemize}
        \item Input dimensions: 1024
        \item Output dimensions: 1024
        \item Layers: 5
        \item Number of heads: 8
        \item Number of indices: 64
    \end{itemize}
    \item Set Transformer Decoder
    \begin{itemize}
        \item Input dimensions: 1024
        \item Hidden dimensions: 1024
        \item Output dimensions: 1024
        \item Layers: 3
        \item Number of heads: 8
    \end{itemize}
    \item Feed-forward Layer
    \begin{itemize}
        \item Hidden dimensions: 2048
        \item Layers: 4
    \end{itemize}
    \item Learning rate: $1 \times 10^{-5}$
    \item Weight decay: $1 \times 10^{-6}$
    \item Optimizer: SGD
    \item Dropout: 0.2
    \item Layer normalization: yes
    \item Training data percentage: 75\%
    \item Test data percentage: 15\%
    \item Validation data percentage: 5\%
\end{itemize}

\section{Algorithm for Applying to Larger Systems}

\renewcommand{\algorithmicrequire}{\textbf{Inputs:}}
\renewcommand{\algorithmicensure}{\textbf{Output:}}
\algnewcommand\algorithmicforeach{\textbf{for each}}
\newcommand{\algorithmicbreak}{\textbf{continue}}
\newcommand{\Continue}{\STATE \algorithmicbreak}
\algdef{S}[FOR]{ForEach}[1]{\algorithmicforeach\ #1\ \algorithmicdo}

\begin{algorithm}
    \caption{Algorithm for Applying to Larger Systems}
    \begin{algorithmic}
        \Require State transitions $S$, Herbrand base $\mathcal{B}$, Trained model $\delta$LFIT2$^n$
        \Ensure Logic program $P$
        \State $\mathcal{T} =$ All subsets of $2^{\mathcal{B}}$ with elements equal to $n$
        \ForEach{$v \in \mathcal{B}$}
        \ForEach{$s \in \mathcal{T}$}
        \State $\mathcal{V} = \Omega(s)$
        \State $S_v = \Omega(S)$
        \If{$S_v$ is not consistent}
        \State \textbf{continue} \Comment{// if the states are not consistent after mapping, skip it}
        \EndIf
        \State $P_v = \delta\text{LFIT2}^n(S_v)$
        \State $P_v = \Omega^{-1}(P_v)$
        \State $P = P \cup P_v$
        \EndFor
        \EndFor
    \end{algorithmic}
\end{algorithm}

$\Omega$ is the mapping function that maps variables into a new assignment of $n$ variables.

\bibliographystyle{splncs04}
\bibliography{references}

\end{document}